\documentclass{article}

\usepackage[nonatbib,preprint]{neurips_2024}

\usepackage[utf8]{inputenc} 
\usepackage[T1]{fontenc}    
\usepackage{hyperref}       
\usepackage{url}            
\usepackage{booktabs}       

\usepackage{colortbl}

\usepackage{url}
\def\redc{\cellcolor[HTML]{FF999A}}
\def\orangec{\cellcolor[HTML]{FFCC99}}
\def\yellowc{\cellcolor[HTML]{FFF8AD}}
\usepackage{amsfonts}       
\usepackage{nicefrac}       
\usepackage{microtype}      
\usepackage{xcolor}
\usepackage{float}

\usepackage{floatflt}
\usepackage{graphicx}
\usepackage{wrapfig}
\usepackage{colortbl}
\usepackage{pifont}
\usepackage{amsmath}
\usepackage{tikz}
\usepackage{enumitem}

\makeatletter
\newcommand{\printfnsymbol}[1]{%
  \textsuperscript{\@fnsymbol{#1}}%
}
\makeatother

\def\R{\mathbb{R}}

\def\G{\mathcal{G}}
\def\ga{\mathcal{T}}

\def\N{\mathcal{N}}
\def\m{\mathrm{m}}

\def\m{{\bf m}}
\def\r{{\bf r}}
\def\v{{\bf v}}

\def\alphab{{\boldsymbol \alpha}}

\def\our{D-MiSo}
\def\ourfull{Dynamic Multi-Gaussian Soup}

\def\redc{\cellcolor[HTML]{FF999A}}
\def\orangec{\cellcolor[HTML]{FFCC99}}
\def\yellowc{\cellcolor[HTML]{FFF8AD}}

\title{
\our{}: Editing Dynamic 3D Scenes using Multi-Gaussians Soup\\
}

\author{%
  Joanna Waczyńska\thanks{Equal contribution} \ 
  \thanks{ \texttt{joanna.waczynska@doctoral.uj.edu.pl}}  
    \\
    Doctoral School of Exact and Natural Sciences\\    
    Jagiellonian University\\
  \And
  Piotr Borycki\printfnsymbol{1}\\
  Faculty of Mathematics and Computer Science\\ 
  Jagiellonian University\\
  \And
   Joanna Kaleta\\
  Warsaw University of Technology \\
  Sano Centre for Computational Medicine \\
  \And
  Sławomir Tadeja\\
  Department of Engineering\\
  University of Cambridge\\ 
  \And
  Przemysław Spurek
  \\
    Jagiellonian University\\ 
    IDEAS NCBR\\
}

\begin{document}

\maketitle
\vspace{-0.8cm}
\begin{abstract}

Over the past years, we have observed an abundance of approaches for modeling dynamic 3D scenes using Gaussian Splatting (GS). Such solutions use GS to represent the scene's structure and the neural network to model dynamics. Such approaches allow fast rendering and extracting each element of such a dynamic scene. However, modifying such objects over time is challenging. SC-GS (Sparse Controlled Gaussian Splatting) enhanced with Deformed Control Points partially solves this issue. However, this approach necessitates selecting elements that need to be kept fixed, as well as centroids that should be adjusted throughout editing. 
Moreover, this task poses additional difficulties regarding the re-productivity of such editing. To address this, we propose  {\bf D}ynamic {\bf M}ult{\bf i}-Gaussian {\bf So}up (\our{})\footnote{\url{https://github.com/waczjoan/D-MiSo}.}, which allows us to model the mesh-inspired representation of dynamic GS. Additionally, we propose a strategy of linking parameterized Gaussian splats, forming a Triangle Soup with the estimated mesh. Consequently, we can separately construct new trajectories for the 3D objects composing the scene. Thus, we can make the scene's dynamic editable over time or while maintaining partial dynamics. 
\end{abstract}


\section{Introduction}
\label{sec:intro}

\begin{wrapfigure}{r}{0.5\textwidth}
\centering
\vspace{-1.5cm}

\includegraphics[width=0.45\textwidth, trim=60 60 60 0, clip]{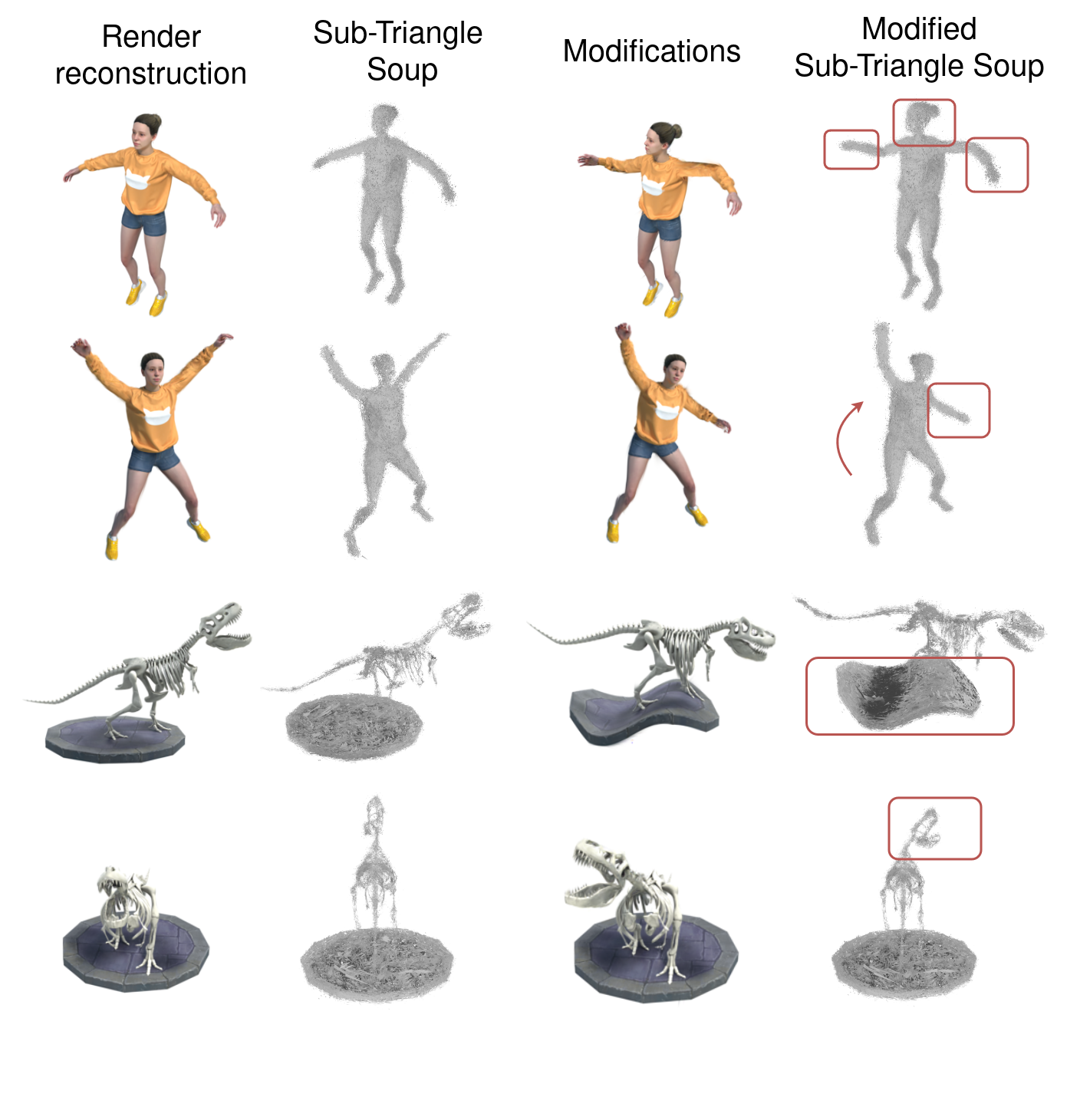}
\caption{\our{} model parameterized dynamic scenes by Triangle Soup (disjoint triangles cloud), which allows modification of objects during time.  }
\label{fig:tesser} 
\vspace{-1.2cm}
\end{wrapfigure}


Recently introduced Gaussian Splatting (GS)~\cite{kerbl20233d} represents the 3D scene structure through Gaussian components. We can combine GS with the neural network (i.e., deform network) to model dynamic scenes \cite{kratimenos2024dynmf}. This approach involves the joint training of both the GS components and the neural network. GS characterizes the 3D object's shape and color, while the neural network utilizes time embedding and Gaussian parameters to generate updated initial positions to model dynamic scenes. Such an approach allows for fast rendering and extracting each element of a dynamic scene. 

\begin{figure}[t]
    \centering
    \includegraphics[width=\textwidth, trim=0 0 0 0, clip]{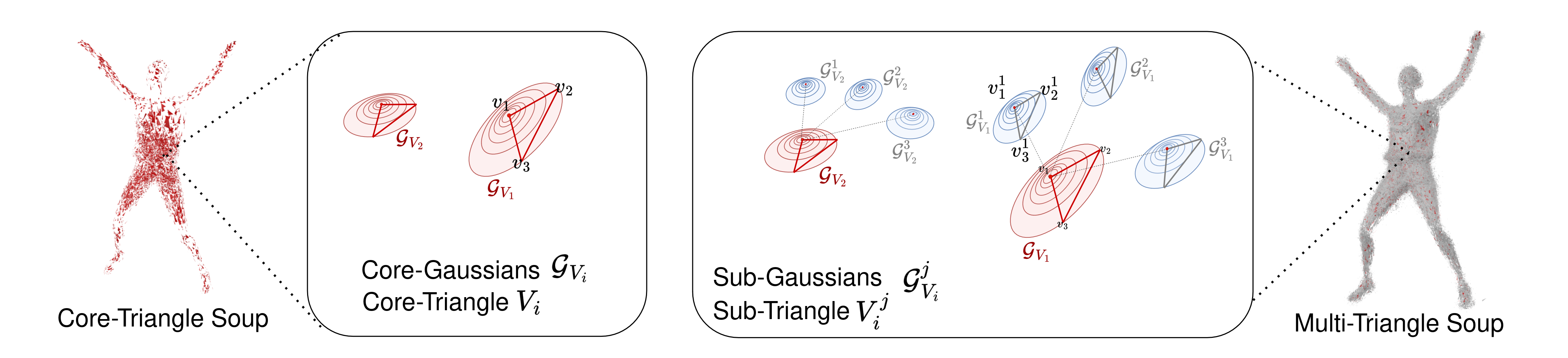}
    \caption{Each object using the \our{} model is represented by Core-Gaussians and Sub-Gaussians, which form Multi-Gaussians. Each Gaussian is related to a triangle using parameterization proposed in GaMeS \cite{waczynska2024games}. Triangles define the Gaussian shape (i.e., location, scale, rotation), and triangles clouds form Triangles Soups.}
\label{fig:coremeshes} 
\vspace{-0.3cm}
\end{figure}

Most existing methods can effectively model dynamic scenes, but generating new 3D objects' positions remains challenging. Consequently, we cannot edit objects over time when using such approaches. To tackle this issue, SC-GS (Sparse Controlled Gaussian Splatting) \cite{huang2023sc} uses Deformed Control Points to manage Gaussians. 
After the training phase, we can manually modify the model at any point in time. However, this method requires identifying elements to remain static and adjusting 3D objects' centroids (nodes) during editing when a relationship between the selected nodes is visible. For example, by moving the humanoid 3D model's hand, the part of the head or leg is also changed. 

To address this issue, we introduce {\bf D}ynamic {\bf M}ult{\bf i}-Gaussian {\bf So}up (\our{}), which is easier to modify (see Fig.~\ref{fig:tesser}) and obtain renders comparable to SC-GS.
\our{} estimates the mesh as Triangle Soup, i.e. a set of disconnected triangle faces \cite{1290060,inproceedings}, and uses a dynamic function to control the vertices.

\begin{wrapfigure}{r}{0.5\textwidth}
\centering
\vspace{-0.7cm}
\includegraphics[width=0.45\textwidth, trim=50 0 50 5, clip]{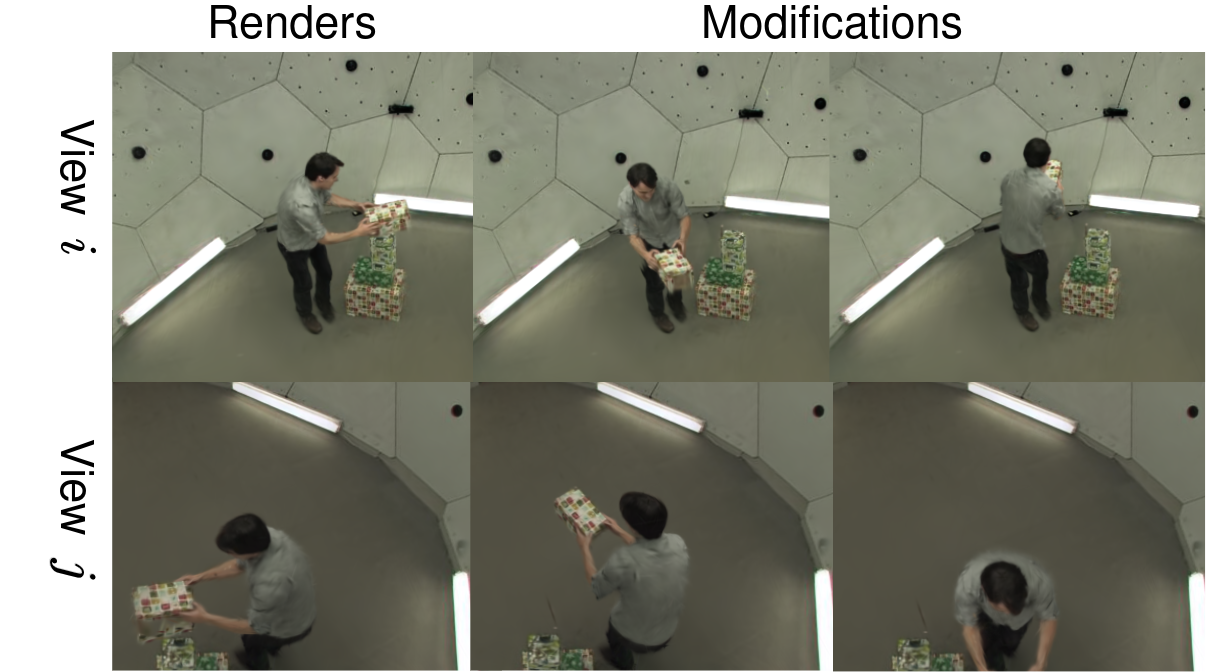}
\caption{\our{} allows us to modify scenes in similar ways as classical mesh-based models.}
\label{fig:boxes}
\end{wrapfigure}

\begin{figure}[ht]
    \centering
    \includegraphics[width=\textwidth, trim=0 0 0 0, clip]{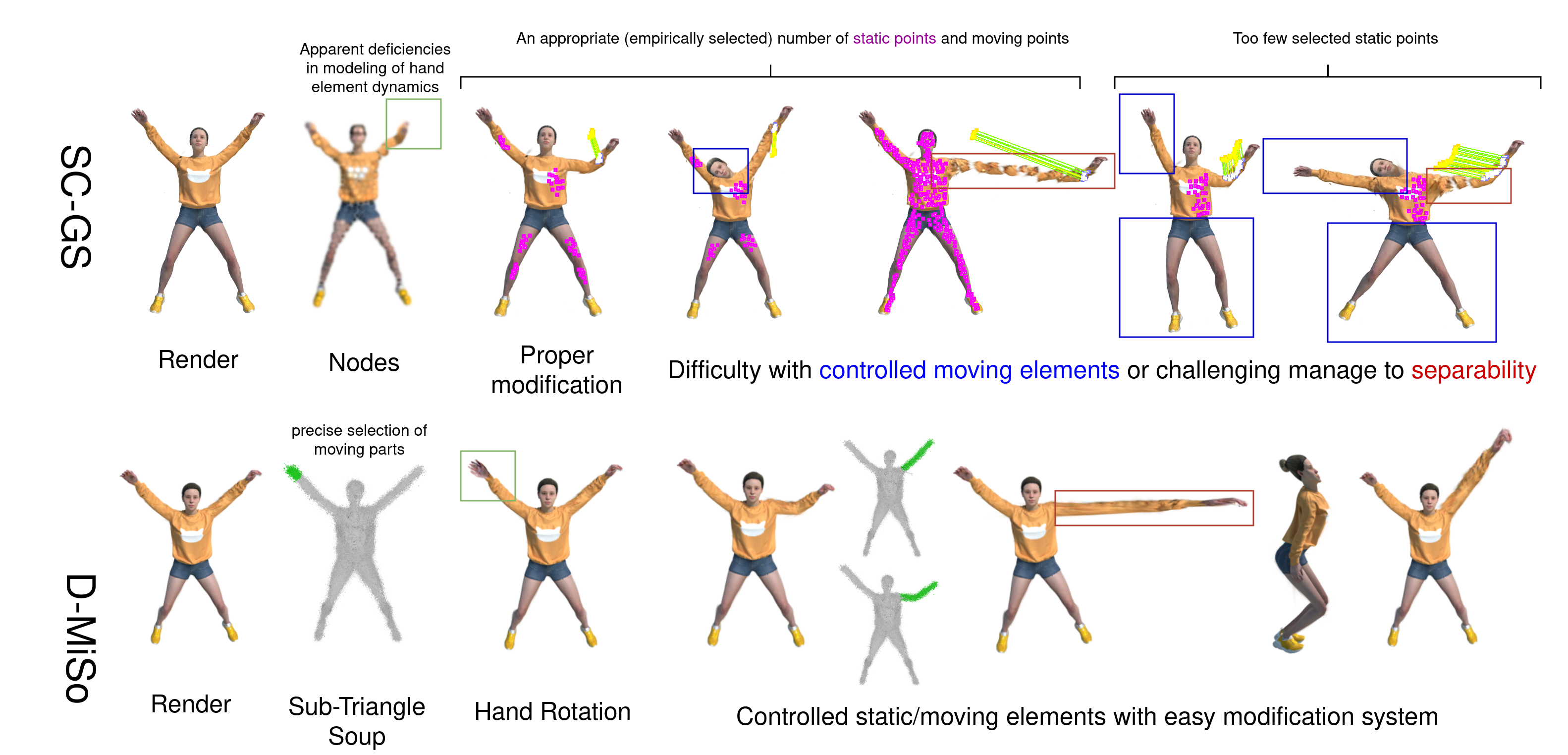}

    \vspace{-0.2cm}
    \caption{Comparison of possible modifications in \our{} and the SC-GS. In the latter, authors use nodes while \our{} apply Sub-Triangle Soup (see the second column). We also must add static (pink) and dynamic (yellow) points in SC-GS to obtain modification by editing dynamic points. In practice, we have to use many static points to stop artifacts. Moreover, SC-GS is not an affine invariant and produces space when we change the size of the objects.
    In the case of \our{}, we marked points and applied modifications. Our model is superior in handling object scaling. 
}
\vspace{-0.5cm}
\label{fig:zmieniajacesiepunkty} 
\end{figure}

Thanks to using Triangle Soup, we control two types of Gaussian components. \our{} employing Multi-Gaussians, defined as larger Core-Gaussians encompassing smaller ones termed Sub-Gaussians (Fig.~\ref{fig:coremeshes}). Sub-Gaussians are defined in the local coordinate system given by principal components of Core-Gaussian. Therefore, by modifying Core-Gaussian, we change all Sub-Gaussians, which allows scene modifications (Fig~\ref{fig:boxes}). Core-Gaussians are an alternative to the control points discussed in \cite{huang2023sc}, with the added advantage of allowing individual modifications. Consequently, there is no necessity for static and dynamic markers. 
In Fig.~\ref{fig:zmieniajacesiepunkty} we present the difference between modification applied by SC-GS and \our{}. In \our{}, we can select and modify one part of the object like a mesh. In contrast, we must select static and dynamic points when using SC-GS, but editing only one part of the object is difficult. 

\begin{figure}[t]
    \centering
    \includegraphics[width=\textwidth, trim=0 0 0 0, clip]{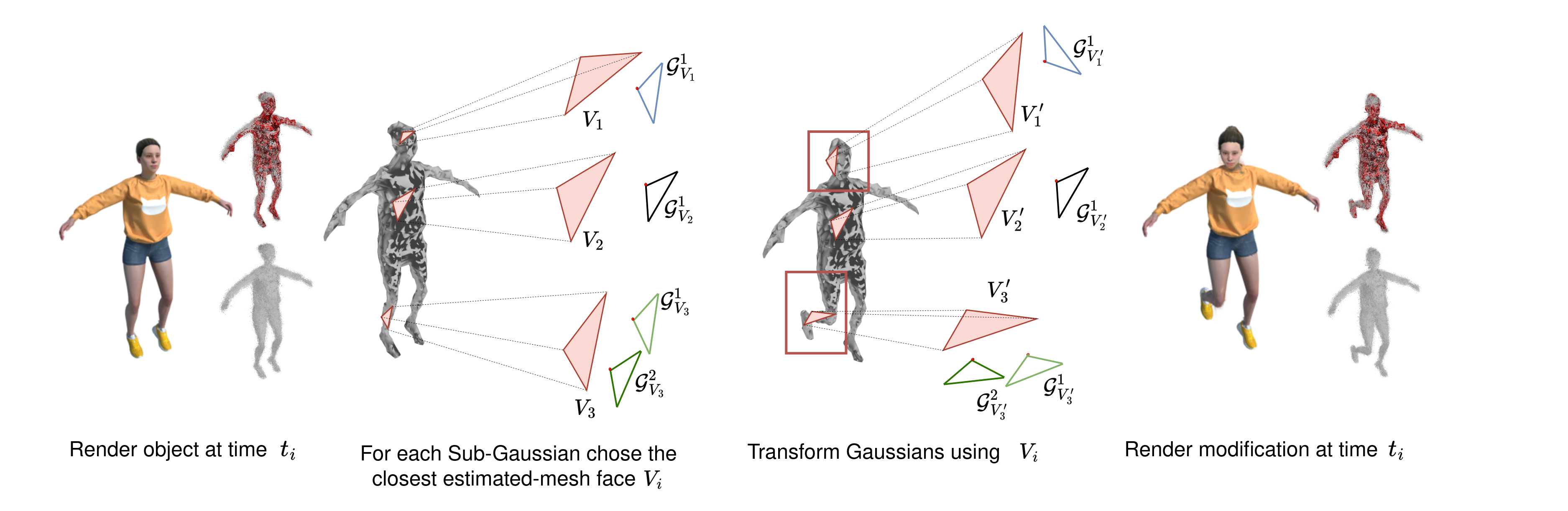}
    \caption{One way to modify the object at the selected time $t_i$ is to take Core-Gaussians and apply a meshing strategy to obtain the correct mesh instead of Triangle Soup. Then, we can parametrize Sub-Gaussian in the coordinate system given bay mesh faces instead of Core-Triangle Soup. Finally, we can modify our mesh to obtain new modifications.
    }
\label{fig:meshandpsedumesh} 
\end{figure}

In \our{}, we use flat Gaussians from GaMeS \cite{waczynska2024games}, which can be parametrized by triangle faces. Consequently, we obtain two Triangle Soups as shown in Fig. \ref{fig:coremeshes} where Core-Triangle Soup is marked by red color, and Sub-Triangle Soup is denoted in blue.
During training, the positions of Core-Gaussians are managed by deformation MLP, while the Sub-Gaussians are collectively manipulated through global transformation and small local deformation. The former describes the general flow of objects in the scene, while the local deformation is responsible for modeling small changes like shadows and light reflections. After training, we can modify our model directly by using the vertex of the Sub-Triangle Soup, or we can generate mesh from the Core-Triangle Soup (Fig. \ref{fig:meshandpsedumesh}). 

The contributions of this paper are significant and are outlined as follows:
\begin{itemize}[leftmargin=0.5cm]
\vspace{-0.2cm}
 \item We introduce the Multi-Gaussian components, which consist of a single large Gaussian response for global transformations and many small components dedicated to rendering. Multi-Gaussian components allow for the modeling of large 3D scene elements.
\vspace{-0.1cm}    
    \item We propose \our{} a model that uses Multi-Gaussian components and two deformation networks for modeling dynamic scenes.
\vspace{-0.1cm}    
    \item Our \our{} allows an object to be edited at a selected moment in time. The edited components are independent, and the editing does not affect other parts of the object. In addition, it also allows for full or partial dynamics to be maintained. Modifications also include scaling and rotation.
\end{itemize}

\section{Related Works}
\vspace{-0.2cm}
Recent advancements in view synthesis, particularly driven by NeRFs \cite{mildenhall2020nerf}, have significantly contributed to the rapid development of novel view synthesis techniques. However, majority of these approaches model static scenes implicitly using multilayer perceptron (MLP). Moreover, several works have extended classical NeRF to dynamic scenes through the use of deformation fields \cite{park2021nerfies,park2021hypernerf,li2021neural} and \cite{pumarola2020dnerf}. The alternative approaches, such as \cite{Gao-ICCV-DynNeRF} and \cite{li-Neural3DVideoSynthesis}, represent scenes as 4D radiance fields. However, NeRF-based solutions often suffer from long training and rendering times. To address this, grid-plane-based methods \cite{Cao2023HexPlane,kplanes_2023,shao2023tensor4d} have been proposed. In addition, several NeRF-based approaches have also been extended for scene editing purposes \cite{yuan2022nerfediting,zheng2023editablenerf,kania2022conerf,Wang2023RIPNeRFLR}.

The recently introduced Gaussian Splatting technique \cite{kerbl20233d} addresses many limitations of other methods, offering multiple advantages due to their explicit geometry representation, enabling easier dynamics modeling. The efficient rendering of 3D GS also avoids densely sampling and querying neural fields, making downstream applications such as free-viewpoint video reconstruction more feasible. While the original GS was developed for static scenes, several extensions for dynamic scenes were proposed. For example, \cite{luiten2023dynamictracking} utilizes a multiview dynamic dataset and a frame-by-frame approach to model such scenes. However, this method lacks inter-frame correlation and requires high storage overhead for long-term sequences. In \cite{yang2023deformable3dgs,kratimenos2024dynmf} MLP is introduced to model changes in Gaussians over time, and in \cite{wu20234dgaussians} MLP together with decomposed neural voxel encoding algorithm are utilized for training and storage efficiency. In \cite{liang2023gaufre}, dynamic scenes are divided into dynamic and static parts, optimized separately and rendered together to achieve decoupling. 
Other works enhance dynamic scene reconstruction using external priors. For example, the diffusion priors can be used as regularization terms during optimization \cite{zhang2024bags}.

Furthermore, GS was employed for mesh-based scene geometry editing. In \cite{gao2024meshbased} 3D Gaussians are defined over an explicit mesh and utilizes mesh rendering to guide adaptive refinement. This approach depends on the extracted mesh as a proxy and fails if the mesh cannot be extracted. In contrast, in \cite{guedon2023sugar} explicit meshes are extracted from 3D GS representations by regularizing Gaussians over surfaces. However, this method involves a costly optimization and refinement pipeline. Another example of \cite{huang2023sc-gs} employs sparse control points for 3D scene dynamics, but this method struggles with intense edit movements and necessitates accurate static node selection. Also, \cite{waczynska2024games} combines GS with mesh extraction. However, it only works for static scenes.
In contrast to all these approaches, we propose a D-MiSo, a mesh-based method specifically designed to handle dynamic scenes. D-MiSo leverages a straightforward pipeline of GS techniques to enable real-time editing of dynamic scenes.

\begin{wrapfigure}{r}{0.5\textwidth}
\vspace{-1.5cm}
\centering
\includegraphics[width=0.45\textwidth, trim=100 90 100 100, clip]{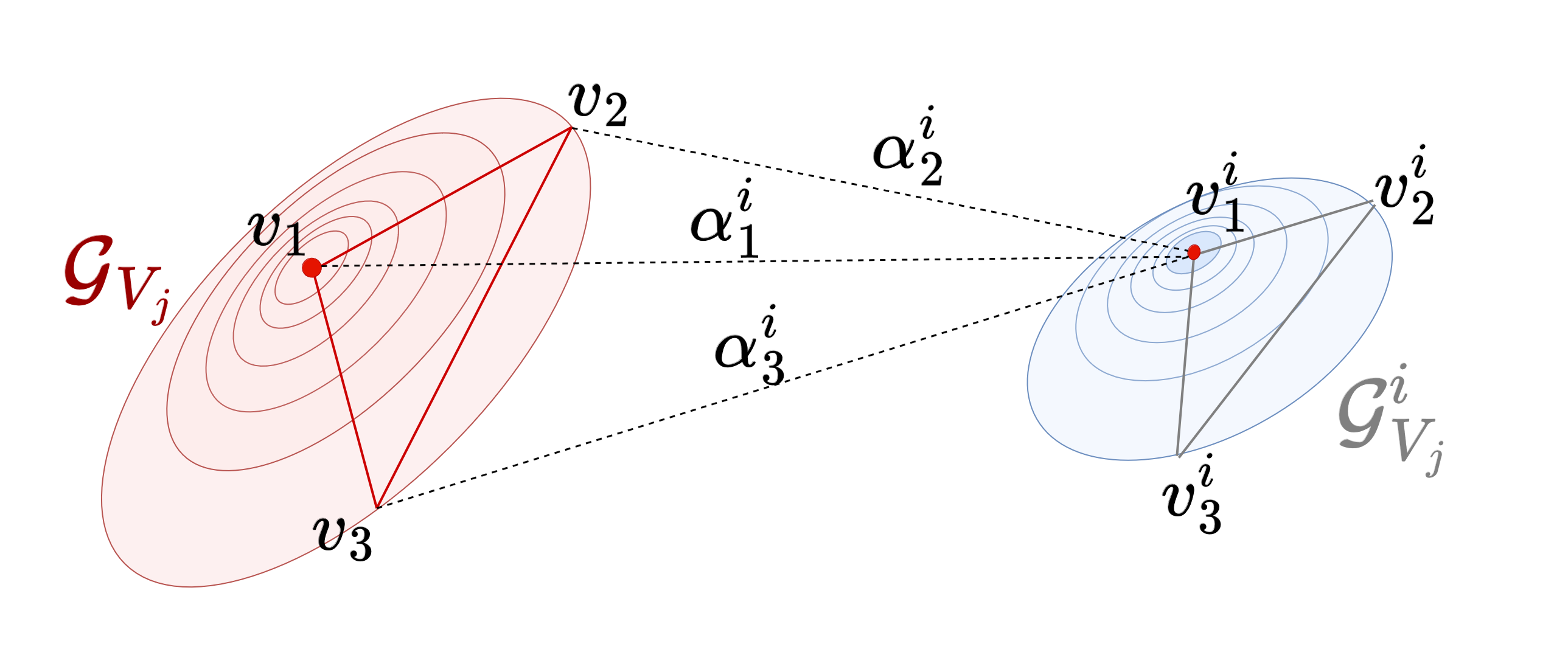}
\caption{Multi-Gaussian, consisting of one Core-Gaussian $\G_{V_j}$ and a Sub-Gaussian $\G_{V_j}^{i}$. The Core Gaussian is parametrized by a $V_{j}$-triangle, and the Sub-Gaussian by a $V_{j}^{i}$-triangle. The relative distance of the center of the Sub-Gaussian from the Core-Gaussian is indicated by $\pmb{\alpha^i} = (\alpha^i_1,\alpha^i_2,\alpha^i_3)$.}
\label{fig:coreand} 
\vspace{-0.5cm}
\end{wrapfigure}

\section{Dynamic Multi-Gaussian Soup}

Here we present the main components of \our{}. We start with the classical GS to provide the foundations for our model. Next, we introduce the concept of Multi-Gaussians and present how to estimate the mesh for editing. Finally, we show \our{}, which uses Multi-Gaussians in dynamic 3D scenes.

\paragraph{Gaussian Splatting}
The Gaussian Splatting (GS) technique models 3D scenes using an array of 3D Gaussians, each specified by its mean position, covariance matrix, opacity, and color expressed using spherical harmonics (SH) \cite{fridovich2022plenoxels,muller2022instant}. The GS algorithm constructs the radiance field by iteratively optimizing the parameters of all Gaussian components. Ultimately, the GS efficiency mainly depends on its rendering method, which involves projecting Gaussian components.

The GS framework employs a dense collection of 3D Gaussians: 
$
\G = \{ (\N(\m_i,\Sigma_i), \sigma_i, c_i) \}_{i=1}^{n},     
$
where $\m_i$ denotes the position, $\Sigma_i$ the covariance, $\sigma_i$ the opacity, and $c_i$ the SH colors for the $i$-th Gaussian.
The GS optimization process involves a repetitive cycle of rendering and comparing the resultant images with the training views. 
In our work, we will use Multi-Gaussian approaches (Fig.~\ref{fig:coreand}).

\paragraph{Multi-Gaussians}

Multi-Gaussians $\G_{multi}$ are dedicated to describing relatively large parts of the 3D scene to allow modification of large blocks instead of modifying each small Gaussian separately. The Multi-Gaussian model comprises a primary large 3D Gaussian (referred to as the Core-Gaussian $\G_{core}$), which encompasses numerous smaller Gaussians (termed Sub-Gaussians $\G_{sub}$), all of which are parameterized by the main Core-Gaussian. 
Multi-Gaussianare is similar to anchor Gaussians from~\cite{yang2024specgaussian}, but we do not use a neural network to produce child components. We parametrize Sub-Gaussians in a local coordinate system.

Similarly to classical GS, we parameterize the Core-Gaussian distribution by center $\m$ and the covariance parameterized by factorization:
$
\Sigma = RSSR^T,
$
where $R$ is the rotation matrix and $S$ the scaling parameters.
More precisely we consider $p$ Core-Gaussians uses flat Guassinas as in \cite{waczynska2024games}, and is defined by:
\begin{equation}\label{eq:core}
\G_{core} = \{(\N_{core}(\m_i,R_i,S_i), \sigma_i, c_i )\}_{i=1}^{p},   
\end{equation}

where $S=\mathrm{diag}(s_1,s_2,s_3),$ $s_1=\varepsilon$ and $R$ is rotation matrix of Core-Gaussian defined as:
$
R= (\r_1,\r_2,\r_3), \mbox{ where } \r_i \in \R^3 
$
which can be interpreted as a local coordinate system used by Sub-Gaussian.

Sub-Gaussian can be interpreted as a child of Core-Gausian. We define centers of Sub-Gaussian $\N_{sub}(\m^{i},R^i,S^i)$ in the local coordinate system of Core-Gaussian 
$\N_{core}(\m,R,S)$ by:
\begin{equation}\label{eq:sublocal}
    \m^i = \m + R\pmb{\alpha^i}^T
\end{equation}
where  $\m$, $R$ is Core-Gaussian position and rotation; and 
$
\pmb{\alpha^i} = (\alpha^i_1,\alpha^i_2,\alpha^i_3)
$
are trainable parameters used to define the positions of the Sub-Gaussian relative to the Core-Gaussian (Fig. \ref{fig:coreand}). 
Sub-Gaussians are used for rendering and can be seen as a main component of our model:
\begin{equation}\label{eq:sub}
\G_{sub} =  \left\{ \left( \N_{sub} \left( \m + R\pmb{\alpha^i}^T ,R^i,S^i\right), \sigma^i, c^i  \right) \right\}_{i=1}^{k},
\end{equation}
where $\m,R,S$ are parameters of Core-Gaussian and, $\alphab^{i},S^i,R^i$ marks parameters of $i$-th Sub-Gaussians with opacity $\sigma^i$, and SH colors $c^i$.


\begin{figure}
    \centering
    \includegraphics[width=\textwidth, trim=0 0 0 0, clip]{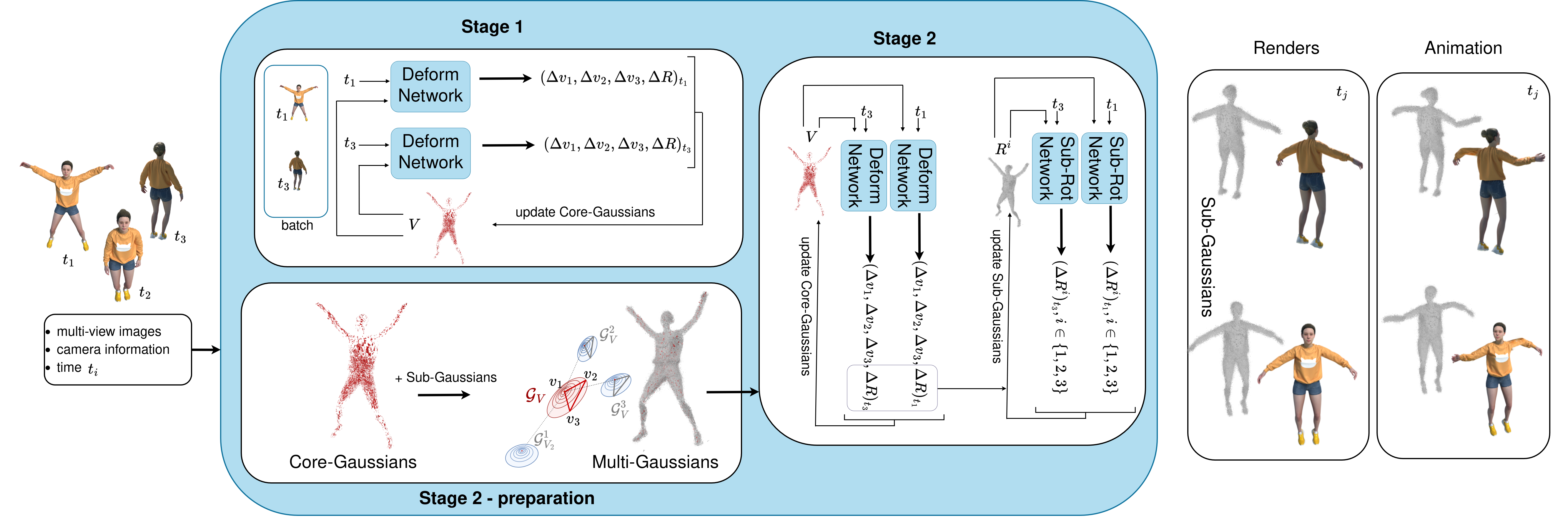}
    \caption{\our{} model diagram. The input of the model consists of images at different moments in time and information regarding the position of the camera. Training distinguishes two main phases (i.e., stages). The first includes the preparation of Core-Gaussians, describing the movement of the object. The second focuses on the fitting of Sub-Gaussians responsible for the render's quality. The model has the ability to produce high-quality renders or create an animation/modification of the object due to Sub-Gaussians (i.e., Sub-Triangles Soup) shape modification.}
\label{fig:algorithm} 
\end{figure}







\begin{wrapfigure}{r}{0.5\textwidth}
\centering
\vspace{-0.5cm}
\centering
    \includegraphics[width=0.45\textwidth, trim=0 0 0 80, clip]{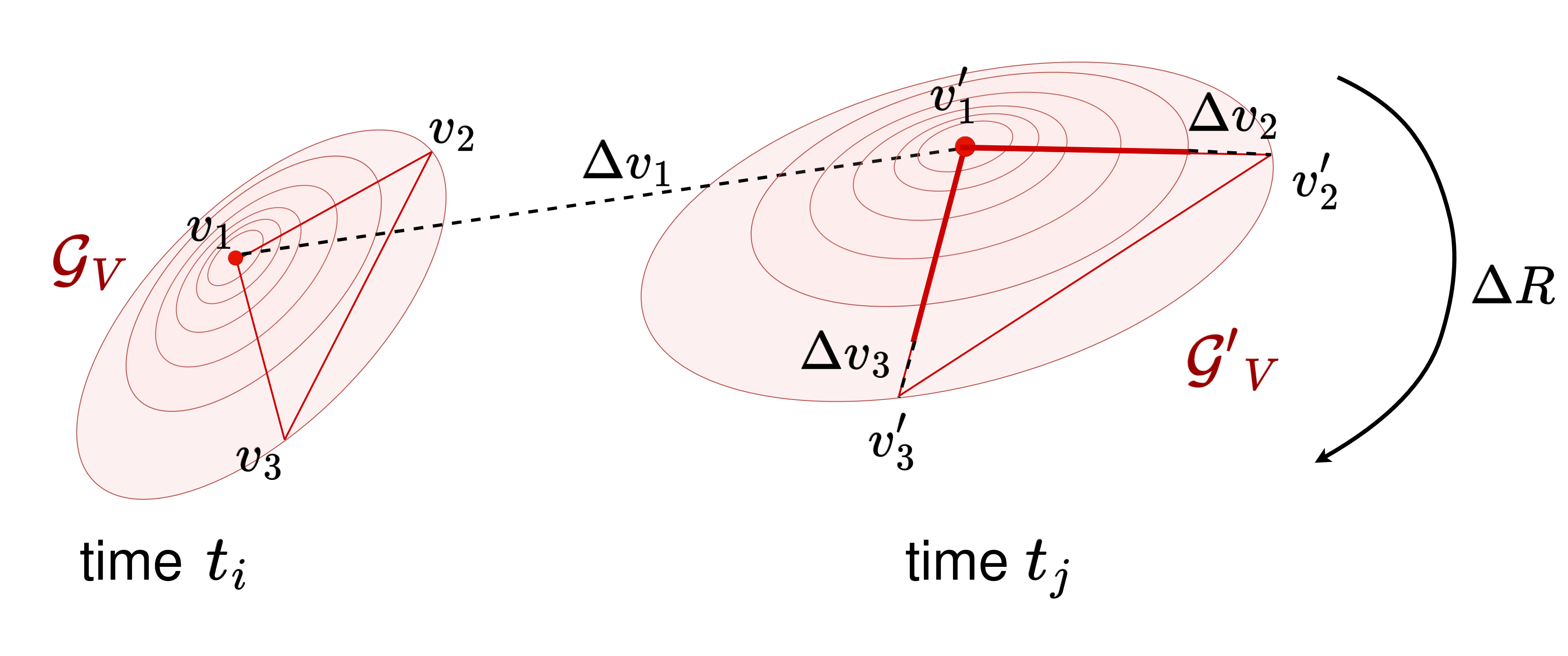}
\caption{Representation of change over time acting on Core-Gaussians using a neural network responsible for movement. In practice, $t_i$ is an abstract time. The network's output returns information about the change in location $\Delta v_1$, scale ($\Delta v_2$, $\Delta v_3$), and  rotation $\Delta R$.}
\label{fig:parametryzujG} 
\end{wrapfigure}

Core-Gaussians is generally dedicated mainly to transformations, hence, in practice opacity or colors are not used during rendering. On the other hand, Sub-Gaussian is devoted to rendering and modification. It has its own opacity and colors. 

One of our mode's most important properties is its ability to model dynamic scenes in each time step. To obtain such properties, we parameterize all Gaussian using Triangle Soup proposed by GaMeS~\cite{waczynska2024games}. Thanks to a few simple transformations, we can convert the mean $\m$ rotation $R$ and scaling $S$ into triangle $V=(\v_1,\v_2,\v_3)$, which parametrize Gaussian distribution. Such transformation is unambiguous and reversible, for more details see Appedix~\ref{app:1}. GaMeS transformation between Gaussian parameters and triangle face we denote by $\ga$:
$$
\N(V) = \N( \ga^{-1}(V) ) =\N(\hat \m_V, \hat R_{V}, \hat S_{V}) 
$$

In our \our{} we parameterize the $p$ Core-Gaussians with $k$ 
Triangle soup:

\begin{equation}\label{eq:mix}
    \G_{multi} =  \left\{\left( \N_{core}(  V_j ) , \left\{ \left( \N_{sub} \left( \hat\m_{V_j} + \hat R_{V_j}\pmb{\alpha^i}^T,\hat R^i,\hat S^i\right), \sigma^i, c^i  \right) \right\}_{i=1}^{k} \right)\right\}_{j=1}^{p},
\end{equation}

where $V_j$, $\pmb{\alpha^i}$,  $R^i$,$\hat S^i$, $\sigma^i$, $c^i$ are trainable parameters and $(\hat \m_{V_j}, \hat R_{V_j}, \hat S_{V_j}) = \ga^{-1} (V_j)$.
Alternatively, we can parameterize Core-Gaussian and Sub Gaussians by Triangle Soup:

\begin{equation}\label{eq:full}
    \G_{multi} =  \left\{\left( \N_{core}(  V_j ) , \left\{ \left( \N_{sub} \left( V_j^i\right), \sigma^i, c^i  \right) \right\}_{i=1}^{k} \right)\right\}_{j=1}^{p},
\end{equation}

where 
$(\hat \m_{V_j}, \hat R_{V_j}, \hat S_{V_j}) = \ga^{-1} (V_j)$, $V^{i} = \ga (\hat\m_{V} + \hat R_{V}\pmb{\alpha^i}^T,\hat R^i,\hat S^i)$.

\begin{wrapfigure}{r}{0.5\textwidth}
\centering
\includegraphics[width=0.45\textwidth, trim=0 0 0 0, clip]{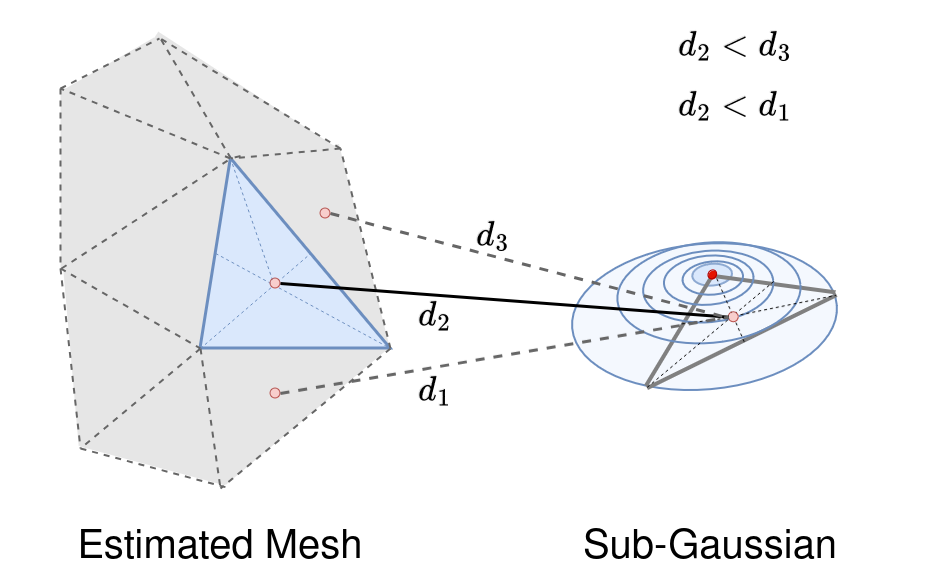}
\caption{One way to animate is to assign the nearest triangle (face) to each Sub-Gaussian from the estimated mesh. The mesh modification changes the assigned Gaussian.}
\label{fig:wyborfaces}
\vspace{-0.7cm}
\end{wrapfigure}

In \our{}, we use the collation of Multi-Gaussian distribution for rendering and Sub-Triangle Soup for editing. The formal definition of our model uses equation \eqref{eq:mix} since in training, we store Core-Gaussian as a triangle face (Core-Gaussian does not have colours) and Sub-Gaussian as a collection from classical GS components with colour and opacity. After training, we parametrize our model to equation \eqref{eq:full} for editing.


%


\subsection{\ourfull{} (\our{})}

\begin{wrapfigure}{r}{0.5\textwidth}
\centering
\vspace{-0.5cm}
\includegraphics[width=0.39\textwidth, trim=40 50 40 5, clip]{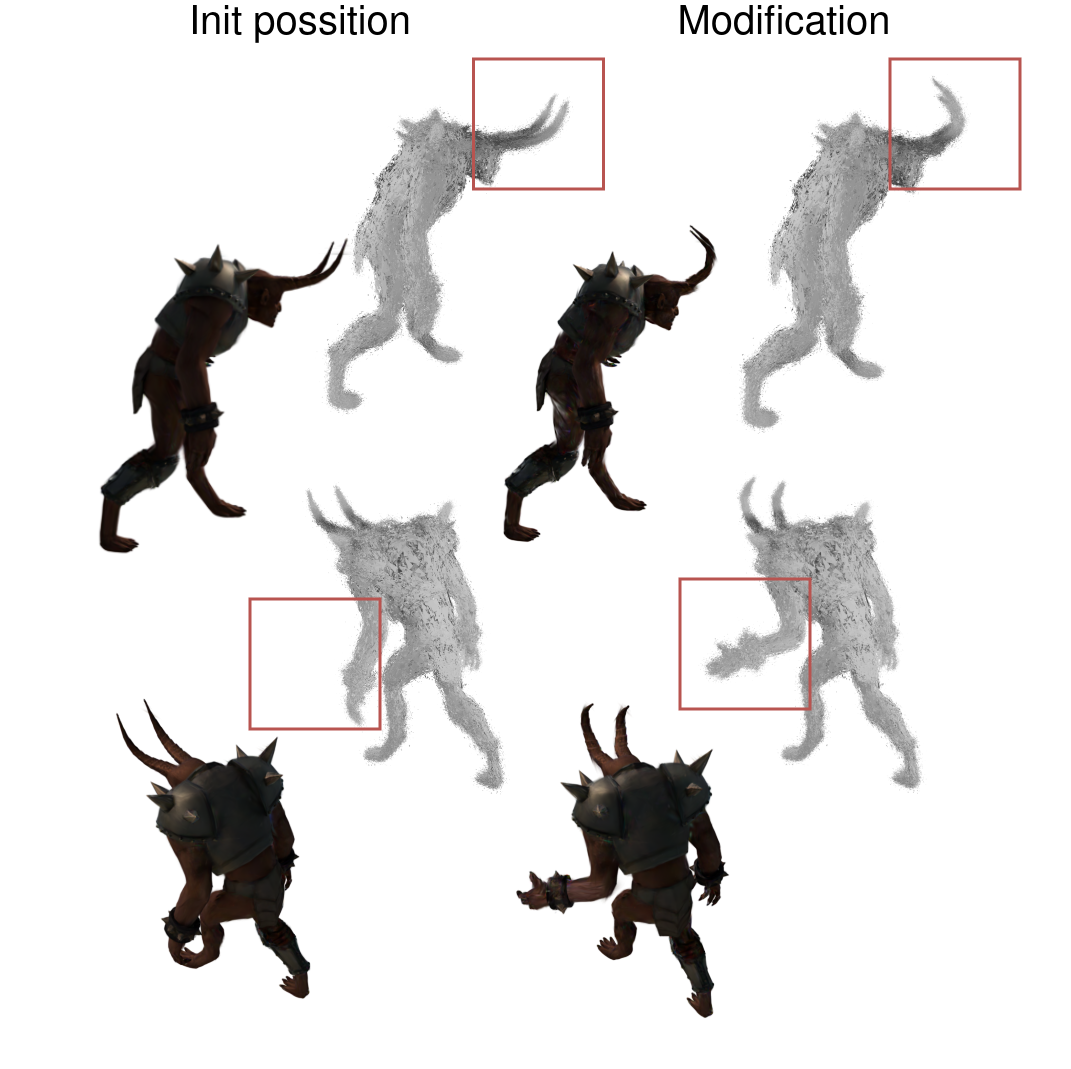}
\caption{An example of Sub-Triangle Soup modification and the render obtained by this change from a different viewpoint. It is possible not only to change the position of the hand but also to raise the thumb.}
\label{fig:hellwarior} 
\vspace{-0.5cm}
\end{wrapfigure}

Previously, we defined Multi-Gaussians and their parametrization using Triangle Soup. 
Now, we have all the tools to present the \our{} model. The overview of our method is illustrated in Fig.~\ref{fig:algorithm}. 
The input to our model is a set of images of a dynamic scene, together with the time label and the corresponding camera poses. Our training is divided into two stages. In the first stage, we initialise the Core-Gaussians. In the second, we add Sub-Gaussian components. 



\paragraph{Stage 1}
First, we train only Core-Gaussians to obtain good initialization for Multi-Gaussins. As Core-Gaussians are mainly employed to capture motion, the model does not require much of them, see Fig.~\ref{fig:algorithm}. In our approach, the Core-Triangle Soup, constructed via the Core-Gaussians parameterization (as depicted in Fig. \ref{fig:coremeshes}), is adjusted depending on the time t.

Practically, when random initialization of Gaussians is necessary, redundant Gaussians must be pruned first to ensure that the remaining ones emphasize the object's shape. To reduce the number of Gaussians and obtain consistent Core-Gaussians, we train GS on a batch containing a few views instead of one. In practice, we render few views from different positions and use back-propagation. 

\begin{figure}
    \centering
    \includegraphics[width=\textwidth, trim=0 0 0 0, clip]{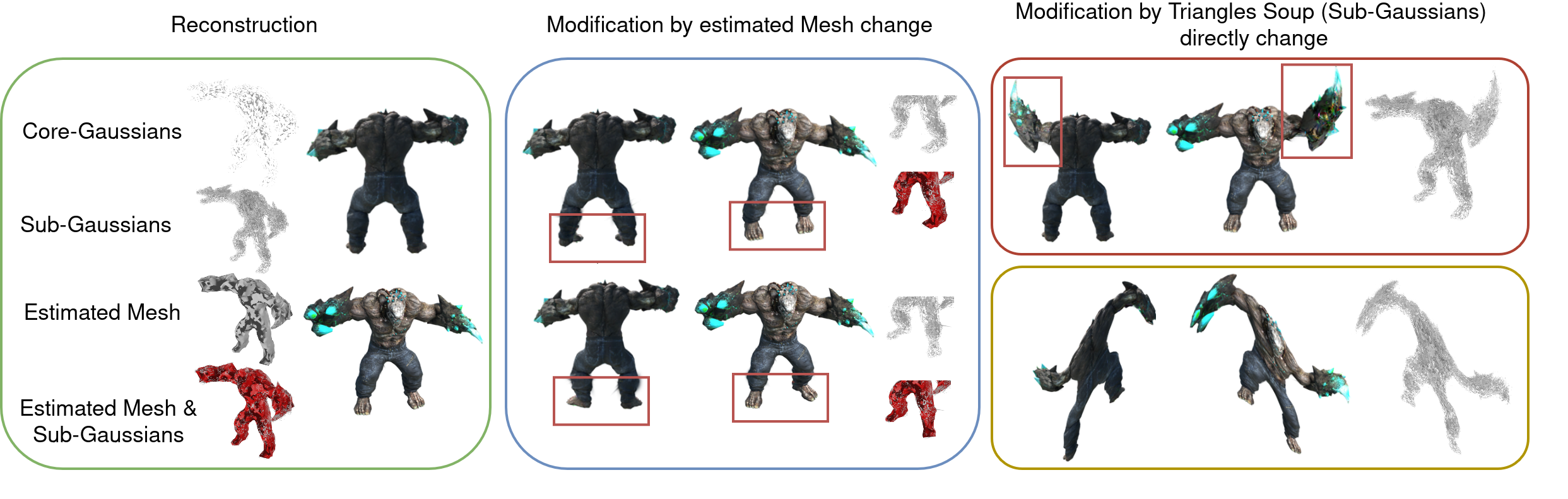}
    \caption{Reconstruction and three ways of modification of the output object. The first involves modifying the estimated mesh, which does not have to be accurate. The next two focus on Sub-Triangle Soup editing. The red box shows the direct modification of the Triangle Soup in a logical way (e.g., raising a hand). The yellow box shows a change in space, i.e., giving fluidity to an object by creating an abstract modification.}
\label{fig:mutant} 
\end{figure}

In particular, we parameterize Gaussians $\N(\m,R,S)$ by face $V=(\v_1, \v_2, \v_3)$ to obtain $\N(\v_1,\v_2,\v_3)$. 
Our \textit{Deform Network} takes as in input the triangle vertixes $V$ assigned to the parameterized 3D Core-Gaussians and the current time $t$ and returns updated
$
\psi(V, t) = (\Delta \v_1(t), \Delta \v_2(t), \Delta \v_3(t), \Delta R_{V}(t)).
$
Such updates consist of translation and rotation (see Fig.~\ref{fig:parametryzujG}):
$$
V(t) = V \odot \psi(V, t) = (\v_1+\Delta \v_1(t), \v_2+\Delta \v_2(t), \v_3+\Delta \v_3 (t))\cdot \Delta R_{V}(t).
$$


\our{} parameterized Core-Gaussians in time $t$ by:
$
\N_{core}(V(t), \sigma, c ) = \N_{core}(V \odot \psi(V, t), \sigma, c )
$
where $\psi$ is a deformable network that moves triangle $V$ according to time $t$. The opacity and colour of the Core-Gaussian are used only in the first stage.

\paragraph{Stage 2: Preparation}

\begin{wrapfigure}{r}{0.5\textwidth}
\centering
\vspace{-0.5cm}
\includegraphics[width=0.45\textwidth, trim=10 9 10 10, clip]{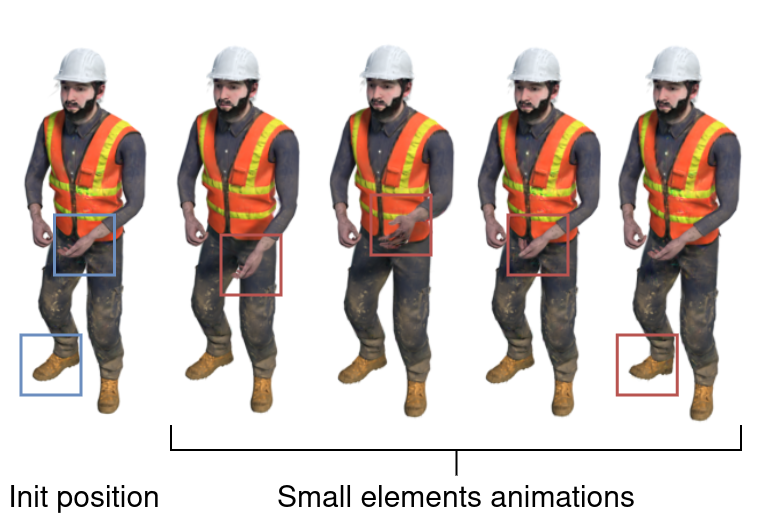}
\caption{Limbs render and modification obtained with \our{}. It is worth noting that it is also possible to close the hand. }
\label{fig:stand} 
\vspace{-1cm}
\end{wrapfigure}

To move to the second phase of the model, it is imperative to prepare the Multi-Gaussians. This involves attaching $k$ Sub-Gaussians to each Core-Gaussian generated in Stage 1, as shown in Fig.~\ref{fig:coreand}. Henceforth, Sub-Gaussians assume responsibility for the resultant rendering. Initially, the Sub-Gaussian adopts the same features as the Core-Gaussian, except the position.


\paragraph{Stage 2}

The primary objective of the second phase is to parallelized Core-Gaussians (Core-Triangles Soup) to enhance understanding of movement and increase rendering quality through the precise training of Sub-Gaussians.
Since the centers of Sub-Gaussians are parameterized by the local coordinate system given by the rotation matrix of Core-Gaussian, when the \textit{Deform Network} $\psi$ changes the Core-Gaussian all Sub-Gaussians (attached to this Core-Gaussian) are modified by global transformation~$\psi(V, t)$.
\our{} use an additional deformation network \textit{Sub-Rot Network} $\phi$ dedicated to each $i$-th Sub-Gassian's small changes. \textit{Sub-Rot Network} takes the Sub-Gaussian rotation matrix $R_V^i$ and the current time $t$ as input and produces an updated rotation matrix $\Delta R_V^{i}(t)$. 

\begin{wrapfigure}{r}{0.5\textwidth}
\centering
\vspace{-0.6cm}
\includegraphics[width=0.13\textwidth, trim=30 60 20 50, clip]{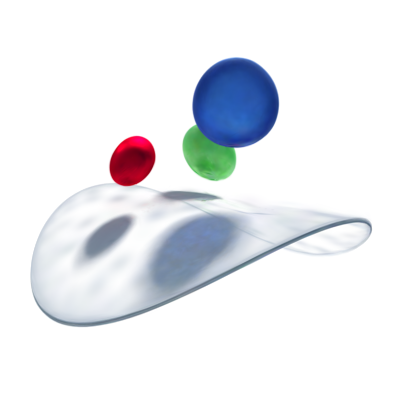} 
\includegraphics[width=0.15\textwidth, trim=30 100 20 50, clip]{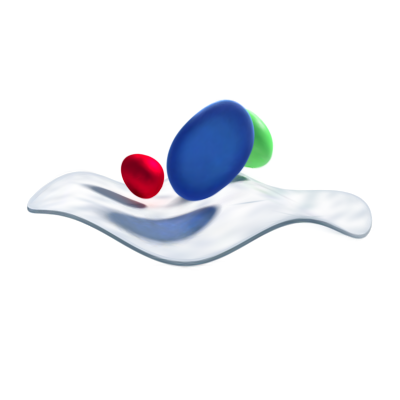}
\includegraphics[width=0.15\textwidth, trim=30 100 20 50, clip]{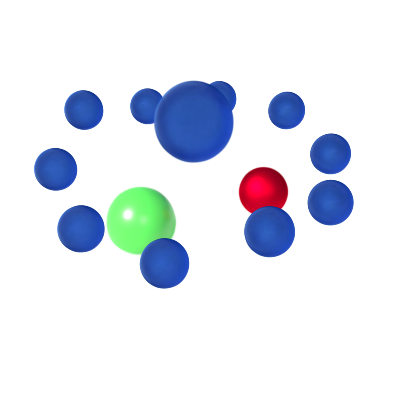}
 \caption{Examples of object modifications. The first method allows for a smooth modification (bending) and also removes (e.g. plate), scales and/or adds (e.g. small blue balls) objects.}
\label{fig:przykladowe_animacje_pilki} 
\end{wrapfigure}

The position $ \hat \m^i_V(t) = \hat \m_{V(t)} + \hat R_{V(t)} \pmb{\alpha^i}^T$ of the Sub-Gaussian in time $t$ is determined by the position $\hat \m_{V(t)}$, and rotation $\hat R_{V(t)}$ of the Core-Gaussian (parameterized by triangle $V$) and the learning parameter $\pmb{\alpha^i}$.
It should be noted that scale $S^i$, color $c_i$, and opacity $\sigma_i$ of Sub-Gaussian are trainable and do not depend on time.
\textit{Sub-Rot Network} produce updated  $\phi( R^i, t ) = \Delta R^i(t)$ for rotation parameter of Sub-Gaussians. 
Finally parameters of Sub-Gaussians in time $t$ depends on \textit{Deform Network} $\psi$, and \textit{Sub-Rot Network} $\phi$, and
the Corr-Gaussians parameter $V$, and Sub-Gaussian parameters $R^i$, $S^i$, $c_i$ and $\sigma_i$:
$$
 \G_{Sub}(t) = \{ (\N_{sub}( \hat \m_{V \odot \psi(V, t)} + \hat R_{V \odot \psi(V, t)} \pmb{\alpha^i}^T ,  R^i  + \phi( R^i, t ), S^i ) , c_i, \sigma_i) \}_{i=1}^{k}.
$$

\our{} final model consists of two deformable networks and two levels of Gaussian distributions. Such a solution allows effectively modeling time in dynamic scenes and modifying objects in each time frame.





The result of the model's inference depends on the camera's view angle as well as on the selected time. Only Sub-Gaussian features are used in the rendering process. Hence, the implementation of generating an output image from Gaussians is no different from a vanilla GS. 

The modifications involve the appropriate transformation of Sub-Gaussians. We base this on the fact that the transformed Gaussians have exactly the same color properties as before editing. Only the shape of the Gaussians is changed. In our work, we present three editing methods (Fig.~\ref{fig:mutant}).

\section{Experiments}

The experimental section is divided into two parts. First, we show that \our{} can model dynamic scenes with high quality; in the second part, that our model allows an easy editing procedure, which is our main contribution.

\subsection{Reconstruction of dynamic scenes}

Here, we outline the specifics of our implementation and provide a detailed description of the used datasets. We demonstrated the core advantages of our model by conducting experiments using three different datasets. The source code is available in GitHub. Our code is developed on top of the GS vanilla code, according to their license. We used NVIDIA GeForce RTX 4090 and A100 GPUs.
The experiments focus on a benchmark tasks of reconstruction. PSNR metric is used to compare our model with another ones. Furthermore, additional numerical comparisons, i.e., SSIM/LPIPS, are available in the supplementary material.

\paragraph{D-NeRF Datasets:} Contains seven dynamic objects with realistic materials described with a single camera~\cite{pumarola2020dnerf}. This means the model had access to only one view at a given moment. Tab.~\ref{tab:psnr} shows we are very comparable to other methods, and we achieve a higher PSNR on one object. The differences in metrics are small, however, our method presents an easier way to edit (Fig.~\ref{fig:zmieniajacesiepunkty}). Providing, among other things, better scaling. 

\begin{wraptable}{r}{7.0cm}
\vspace{-0.5cm}
    \caption{Quantitative comparisons (PSNR) on a D-NeRF dataset showing that \our{} gives comparable results with other models.}
{\scriptsize
\begin{center}
    \begin{tabular}{@{}llllllll@{}}
    \multicolumn{8}{c}{PSNR $\uparrow$} \\
         & Hook & Jumpin. & Trex & Bounc.  & Hell.  & Mutant & Standup \\ 
 \hline
D-NeRF \cite{pumarola2020dnerf} &  29.25 & 32.80  & 31.75  & 38.93  & 25.02  & 31.29  & 32.79 \\ 
TiNeuVox-B \cite{Fang_2022}& 31.45 & 34.23 & 32.70 & 40.73  & 28.17  & 33.61 & 35.43 \\ 
Tensor4D \cite{shao2023tensor4d}  &  29.03 & 24.01& 23.51 & 25.36  &  31.40   & 29.99 & 30.86  \\ 
K-Planes \cite{kplanes_2023}   &  28.59 &  32.27 & 31.41 & 40.61  & 25.27  & 33.79 & 34.31  \\  
FF-NVS \cite{Guo_2023_ICCV}    &  32.29 & 33.55 & 30.71 & 40.02  & 27.71   &  34.97 & 36.91  \\ 
4D-GS \cite{wu20234dgaussians}     &  30.99 & 33.59 & 32.16 & 38.59  & 31.39  &  35.98 & 35.37   \\ 
DynMF \cite{kratimenos2024dynmf}    & \yellowc 33.94 & \yellowc 38.04 & \yellowc35.82 & \orangec 41.92  & \yellowc  37.51  & \yellowc 41.68 & \yellowc 41.16  \\ \hline
        \multicolumn{8}{c}{ Editable} \\ \hline
SC-GS \cite{huang2023sc-gs} & \redc 39.87 & \orangec 41.13 & \redc 41.24 & \redc 44.91  & \redc 42.93  & \redc 45.19 & \redc 47.89  \\ 
\textbf{\our{} (our)}& \orangec 38.13 & \redc 42.05 & \orangec 40.88 & \yellowc 41.49  & \orangec  41.49  & \orangec 44.38 & \orangec 47.66 \\ 
\hline
    \end{tabular}
    \end{center}
    }
\vspace{-0.7cm}    

    \label{tab:psnr}
\end{wraptable}

\paragraph{NeRF-DS\cite{yan2023nerfds}:} This dataset contains again seven real-world scenarios containing a moving or deforming specular object. Each scene was recorded using two cameras, with the video captured by one of them being used as a training set, while the footage from the second one was treated as a test set. The camera pose was estimated using COLMAP for both cameras. Tab. \ref{tab:psnr_nerfds} shows that our method for four objects achieves the SOTA results.

\paragraph{PanopticSports Datasets:} The dataset comprises six dynamic scenes featuring significant object and actor movements \cite{luiten2023dynamictracking}. The scenes are categorized by the activities performed in the video sequences, i.e., juggling, boxing, softball, tennis, football, and basketball. Each scene was recorded using 31 cameras over 150 timesteps. Following the official data split, we use footage from 27 cameras for training and the remaining 4 cameras for testing. Numerical results are shown in Tab. \ref{tab:panaptic}. The results show that the model achieved for five objects SOTA according to PSNR metrics, and six times according to LPIPS metrics.

We show that \our{} is comparable to other methods. However, our main contribution in comparison to other methods is the very easy editing of the resulting object~(Fig.~\ref{fig:zmieniajacesiepunkty}~\ref{fig:mutant}).

\subsection{Editing of dynamic scenes}



As mentioned earlier, we proposed three new methods of output object modification. The first one focuses on moving and preparing a formal mesh, which is connected in contrast to Triangle Soup. We apply a simple meshing strategy on Core-Gaussians. 
We used the basic Alpha Shape algorithm~\cite{1056714,articleEdelsbrunner} and showed that the mesh does not have to estimate the surface perfectly to be effective.
Then, we reparametrize the Sub-Gaussian by finding the closest face from the mesh~(Fig.~\ref{fig:wyborfaces}). In practice, we represent each Sub-Gaussian in a local coordinate system (analogically to Multi-Gaussians). Therefore, each Sub-Gaussian is assigned to the nearest face (Fig. \ref{fig:wyborfaces}). Each face can have a different number of Sub-Gaussians prescribed in such a configuration. This method allows us to maintain the consistency bestowed by the mesh. The whole process is shown in Fig.~\ref{fig:meshandpsedumesh}. 
Thanks to such representation, we can edit our connected mesh to produce the correct edition of the dynamic scene (Fig. \ref{fig:coremeshes}).

\begin{wraptable}{r}{6.cm}
\vspace{-0.7cm}
    \caption{PSNR comparisons on a NeRF-DS dataset showing that \our{} gives comparable results with other models.}
{\scriptsize 
\begin{center}
    \begin{tabular}{@{}llllllll@{}}
    \multicolumn{8}{c}{PSNR $\uparrow$} \\
         & Bell & Sheet & Press & Basin & Cup & Sieve & Plate \\ 
 \hline
HyperNeRF \cite{park2021hypernerf} & \yellowc 24.0 & 24.3 & 25.4 & \orangec 20.2 & \orangec 20.5 & 25.0 & 18.1   \\
NeRF-DS \cite{yan2023nerfds} & 23.3 & \yellowc 25.7 & \orangec 26.4 & \yellowc 20.3 & \redc 24.5 & \orangec 26.1 & \redc 20.8 \\
TiNeuVox-B \cite{Fang_2022} & 23.1 & 21.1 & 24.1 & \redc 20.7 & \orangec 20.5 & 20.1 & \orangec 20.6 \\
\hline
\multicolumn{8}{c}{ Editable  } \\ 
\hline
SC-GS \cite{huang2023sc} & \orangec 25.1 & \redc 26.2 & \redc 26.6 & 19.6 & \redc 24.5 & \yellowc 26.0 & \yellowc 20.2 \\
\textbf{\our{} (our)} & \redc 25.3 & \orangec 25.8 & \yellowc 25.6 & 19.8 & \redc 24.5 & \redc 26.5 & \redc 20.8 \\ 
\hline
    \end{tabular}
    \end{center}
    }
    \label{tab:psnr_nerfds}
\vspace{-0.4cm}
\end{wraptable}

The second editing method allows us to define the connections, i.e., editing Sub-Triangles Soup directly e.g. moving a hand or bending horns (Fig.~\ref{fig:hellwarior}) or rotating a human body (Fig. \ref{fig:boxes}). Changes are possible because we can also transform a group of Sub-Triangles instead of individual ones. This way allows for even very subtle changes like raising the thumb (Fig. \ref{fig:hellwarior}), turning the hand over~(Fig~\ref{fig:zmieniajacesiepunkty}), opening or closing the hand (Fig.~\ref{fig:stand}). These edits would be difficult in other approaches based on adjusting the 3D objects' centroids (nodes) since the space of nodes is limited in details area (Fig. \ref{fig:zmieniajacesiepunkty}). Nodes use is preeminent for defining movement, which was insignificant in these places.

With this method, we can also change complicated objects like a $360^\circ$ scene without losing the dynamics-related model. For example, the rotation of a person stacking boxes (Fig. \ref{fig:boxes}). 

\begin{wraptable}{l}{8.cm}
\vspace{-0.5cm}
\caption{Comparison on PanopticSports dataset.} 
\scriptsize
\begin{tabular}{@{}l@{\;}|cc@{}c|c@{}c@{}c|c@{}c@{}c|c@{}c@{}c@{}}
\toprule
Metrics & Method & Juggle & Boxes & Softball & Tennis & Football & Basketball & Mean   \\
\toprule
 & 3DGS \cite{kerbl20233d} & \yellowc 28.19 & \yellowc  28.74 & \redc 28.77 & \yellowc 28.03 & \orangec 28.49 & \yellowc 27.02 & \yellowc 28.21 \\
PSNR $\uparrow$ & Dyn3DG \cite{luiten2023dynamictracking} & \orangec 29.48 & \redc 29.46 & \yellowc 28.43& \orangec 28.11 & \orangec 28.49 & \orangec 28.22 & \orangec 28.7  \\
& \our{} (our)& \redc 29.79 & \orangec 29.39 & \orangec 28.6& \redc 29.02& \redc 28.99 & \redc 28.49 & \redc 29.04 \\
\midrule
 &3DGS \cite{kerbl20233d} & \yellowc 0.91 & \orangec 0.91 & \orangec 0.91 & \yellowc  0.90 & \yellowc 0.90 & \yellowc 0.89 & \yellowc 
 0.90  \\
SSIM  $\uparrow$ & Dyn3DG \cite{luiten2023dynamictracking} & \orangec 0.92 & \orangec 0.91 & \orangec 0.91 & \orangec 0.91 & \orangec 0.91 & \orangec 0.91 & \orangec 0.91  \\
& \our{} (our)  & \redc 0.93 & \redc 0.92 & \redc 0.92 & \redc 0.92 & \redc 0.92 & \redc 0.91 & \redc 0.92\\
\midrule
 &3DGS \cite{kerbl20233d} & \orangec 0.15   & \orangec 0.15  & \redc 0.14     & \orangec 0.16   & \orangec 0.16 &  \orangec 0.18 & \orangec 0.16  \\
LPIPS $\downarrow$ & Dyn3DG \cite{luiten2023dynamictracking} & \orangec 0.15 & \yellowc 0.17 & \yellowc  0.19 & \yellowc  0.17 & \yellowc  0.19 & \orangec 0.18 & \yellowc  0.17 \\
& \our{} (our) & \redc 0.13 & \redc 0.13  & \orangec 0.15 & \redc 0.14 & \redc 0.13 & \redc 0.15 & \redc 0.14 \\
\bottomrule

\end{tabular}\label{tab:panaptic}
\end{wraptable}

The third method focuses on transforming the space in which the object is located. Similar to the second method, the third one also works directly on the Sub-Triangle soup. We can achieve such an effect by applying a certain function to the selected plane. In practice, the object, or a portion of it, is modified, for instance, through the use of sinusoidal functions. It allows us to obtain fluidity and more easily define the physical nature of the movement (Fig.~\ref{fig:tesser},~\ref{fig:mutant},~\ref{fig:przykladowe_animacje_pilki}).

Our methods are also scalable, so we can easily remove or duplicate elements from an image. Moreover, the duplicated elements can be given their own dynamics. Examples of these effects are shown in Fig. \ref{fig:przykladowe_animacje_pilki} where we removed the plate, and both duplicated and rescaled the blue balls multiple times.

\section{Conclusion}

\our{} is a novel method based on Gaussian Splatting parameterization, which produces a cloud of triangles called Triangle Soup. The method allows easy editing of objects created in inference with the possible transformations including moving, scaling, and rotating. By defining Multi-Gaussians, the obligatory separability of modified elements seen in other models is combated. In addition, certain elements of objects can be duplicated and removed (Fig. \ref{fig:przykladowe_animacje_pilki}). Furthermore, the \our{} method can facilitate giving different dynamics to separate parts of an object (Fig. \ref{fig:tesser}). 

\paragraph{Limitation} 
The method allows for complex changes at a given moment in time. However, if some area is not well represented in the training set, it is not possible to edit them. For example, a person's hand can be changed but not fingers (Fig. \ref{fig:zmieniajacesiepunkty}). This is due to the liminality of Triangle Soup relative to a well-fitted mesh.

\section*{}

\bibliographystyle{unsrt}


\appendix

\section{GaMeS \cite{waczynska2024games} parametrisation of Gaussian component }
\label{app:1}

Multi-Gaussians describe 3D scenes using solid blocks rather than tiny Gaussian distributions. This method enhances our model's suitability for dynamic environments. Additionally, we desire the capability to modify 3D objects at any moment. As a result, we employ a mesh-inspired to condition our model. In order to do that, we incorporate the parametrization concept from GaMeS~\cite{waczynska2024games}.

As it was mentioned earlier, our model uses flat Gaussians. Therefore, we can approximate Gaussian components using triangle face mesh by parameterizing Gaussian components by the vertices of the mesh face. We denote such transformation by $\ga(\cdot)$. In practice as a consequence of parameterizing each Gaussian, we obtain a cloud of triangles called Triangle Soup~\cite{1290060}. Accordingly, we get: Sub-Triangle Soup, Core-Triangle Soup and Multi-Triangle Soup, see Fig.~\ref{fig:coremeshes}.
 
Let us assume that we have a Gaussian component parameterized by mean $\m$, rotation matrix $R=[\r_1,\r_2,\r_3]$ and scaling $S = \mathrm{dig}(\varepsilon,s_2,s_3).$
We define three vertex of triangle (face):
$$
V=[\v_1,\v_2,\v_3],
$$
where $\v_1 = \m$, $\v_2 = \m+s_2  \r_2$, $\v_3 =\m+s_3\r_3$.
 
Now we froze the vertex of the face $V=[\v_1,\v_2,\v_3]$ and reparameterize the Gaussian component by defining $\hat \m$, $\hat R=[\hat \r_1,\hat \r_2,\hat \r_3]$ and
$\hat S = \mathrm{dig}(\hat s_1,\hat s_2,\hat s_3).$
First, we put $\hat \m = \v_1$. 
The first vertex of $\hat R$ is given by a normal vector:
$$
\hat \r_1 = \frac{(\v_2 - \v_1) \times (\v_3 - \v_1)}{\|(\v_2 - \v_1) \times (\v_3 - \v_1) \|},
$$
where $\times$ is the cross product.
The second one is defined by
$\hat \r_2 = \frac{(\v_2-\v_1)}{\|(\v_2-\v_1)\|}.$
The third one is obtained as a single step in the Gram–Schmidt process~\cite{bjorck1994numerics}: 
$$
\hat \r_3 = \mathrm{orth}(\v_3-\v_1;\r_1,\r_2).
$$

Scaling parameters can also be easily calculated as
$s_1= \varepsilon$, $\hat s_2 = \|\v_2-\v_1\|$ and  $\hat s_3 = \langle \v_3-\v_1, \hat \r_3 \rangle$.

Consequently, the covariance of Gaussian distribution positioned to face is given by:
$$
\hat \Sigma_{V} = \hat R_V \hat S_V \hat S_V \hat R_V^T,
$$ 
and correspond with the shape of a triangle $V$. 
For one face~$(\v_1,\v_2,\v_3)$, we define the corresponding Gaussian component:
$$
\N((\v_1,\v_2,\v_3)) =  \N(\hat \m_V, \hat R_{V}, \hat S_{V}) .  
$$
Finally, the Gaussian component is derived from the mesh face parameters. This approach can be applied in a Multi-Gaussian framework.
Therefor we will use invertible transformation between Gaussian parameters and triangle face $T$ and notation
$$
\N(V) := \N( \ga^{-1}(V) ) =\N(\hat \m_V, \hat R_{V}, \hat S_{V}). 
$$

\section{Extension of numerical results from main paper}

This section contains a numerical comparison of the \our{} with others models, regarding the experiments described in the main document. We used additional LPIPS and SSIM metrics. We see that we get comparable competitive results. We can see that our method, because of the LPIPS metric, achieves better results for five objects from D-NeRF dataset (Tab. \ref{tab:apend1}). Numerical comparison with metrics for the NeRF-DS dataset show that we are comparable to other methods in the reproduction task ( Tab. \ref{tab:apend2}).

In our method, a batch of images is taken as an input to the model. Table \ref{fig:apend3} shows a numerical comparison using batch: 4, 8, 10 on D-NeRF. Training takes 80 thousand iterations and second stages started at the 5 thousandth iteration. Each Core-Gaussian have attached 25 Sub-Gaussians. This is comparable to the SC-GS implementation. Similar study have been done for NeRF-DS dataset presented in \ref{fig:apend4}. We can see that batch is playing a bigger role in the dataset, which is considering a more pronounced move.

\begin{table}[]
{\small
\begin{center}
    \begin{tabular}{@{}l@{\;}|cc@{}c|c@{}c@{}c|c@{}c@{}c|c@{}c@{}c@{}}
        
        \hline
        Methods & \multicolumn{3}{c}{Hook} & \multicolumn{3}{c}{Jumpingjacks} & \multicolumn{3}{c}{Trex} & \multicolumn{3}{c}{BouncingBalls}\\
        & PSNR $\uparrow$ & SSIM $\uparrow$ & LPIPS $\downarrow$ & PSNR $\uparrow$ & SSIM $\uparrow$ & LPIPS $\downarrow$ & PSNR $\uparrow$ & SSIM $\uparrow$ & LPIPS $\downarrow$ & PSNR $\uparrow$ & SSIM $\uparrow$ & LPIPS $\downarrow$
        \\\hline
        D-NeRF & 29.25 & .968 & .1120 
               & 32.80 & .981 & .0381 
               & 31.75 & .974 & .0367
               & 38.93 & .987 & .1074
        \\
        TiNeuVox-B & 31.45 & .971 & .0569 
                   & \yellowc 34.23 & .986 & .0383
                   & \yellowc 32.70 & .987 & .0340
                   & \yellowc 40.73 & .991 & .0472
        \\
        Tensor4D & 29.03 & .955 & .0499 
                 & 24.01 & .919 & .0768 
                 & 23.51 & .934 & .0640
                 & 25.36 & .961 & .0411
        \\
        K-Planes & 28.59 & .953 & .0581 
                 & 32.27 & .971 & .0389 
                 & 31.41 & .980 & .0234
                 & 40.61 & .991 & .0297
        \\
        FF-NVS & \yellowc 32.29 & \yellowc .980 & .0400 
               & 33.55 & .980 & .0300 
               & 30.71 & .960 & .0400
               & 40.02 & .990 & .0400
        \\
        4D-GS & 30.99 & \orangec .990 & \yellowc .0248
              & 33.59 & \yellowc .990 & \yellowc .0242 
              & 32.16 & \yellowc .988 & \yellowc .0216
              & 38.59 & \yellowc .993 & \yellowc .0267
        \\
        SC-GS & \redc 39.87 & \redc .997 & \redc .0076 
              & \orangec 41.13 & \redc .998 & \orangec .0067 
              & \redc 41.24 & \redc .998 & \orangec .0046
              & \redc 44.91 & \redc .998 & \orangec .0166
        \\
        \our{} (our) & \orangec 38.13 & \orangec .990 & \orangec .0086
               & \redc 42.05 & \orangec .995 & \redc .0049
               & \orangec 40.88 & \orangec .996 & \redc .0029
               & \orangec 41.49 & \orangec .993 & \redc .0079
        \\

        \hline
        Methods & \multicolumn{3}{c}{Hellwarrior} & \multicolumn{3}{c}{Mutant} & \multicolumn{3}{c}{Standup} & \multicolumn{3}{c}{Average} \\
        & PSNR $\uparrow$ & SSIM $\uparrow$ & LPIPS $\downarrow$ & PSNR $\uparrow$ & SSIM $\uparrow$ & LPIPS $\downarrow$ & PSNR $\uparrow$ & SSIM $\uparrow$ & LPIPS $\downarrow$ & PSNR $\uparrow$ & SSIM $\uparrow$ & LPIPS $\downarrow$
        \\\hline
        D-NeRF & 25.02 & .955 & .0633 
               & 31.29 & .978 & .0212
               & 32.79 & .991 & .0241
               & 31.69 & .975 & .0575
        \\
        TiNeuVox-B & 28.17 & \yellowc .978 & .0706 
                   & 33.61 & .982 & .0388
                   & 35.43 & .991 & .0230
                   & 33.76 & .983 & .0441
        \\
        Tensor4D & \yellowc 31.40 & .925 & .0675 
                 & 29.99 & .951 & .0422
                 & 30.86 & .964 & .0214
                 & 27.62 & .947 & .0471
        \\
        K-Planes & 25.27 & .948 & .0775 
                 & 33.79 & .982 & .0207
                 & 34.31 & .984 & .0194
                 & 32.32 & .973 & .0382
        \\
        FF-NVS & 27.71 & .970 & .0500 
               & 34.97 & .980 & .0300
               & \yellowc 36.91 & .990 & .0200
               & 33.73 & .979 & .0357
        \\
        4D-GS & 31.39 & .974 & \yellowc .0436 
              & \yellowc 35.98 & \yellowc .996 & \yellowc .0120
              & 35.37 & .994 & \yellowc .0136
              & \yellowc34.01 & \yellowc .987 & \yellowc .0316
        \\
        SC-GS & \redc 42.93 & \redc.994 & \redc .0155
              & \redc 45.19 & \redc .999 & \orangec .0028
              & \redc 47.89 & \redc .999 & \orangec .0023
              & \redc 43.31 & \redc .997 & \redc .0063
        \\
        \our{} (our) & \orangec 41.49 & \orangec .986 & \orangec .0173
               & \orangec 44.38 & \orangec .997 & \redc .0026
               & \orangec 47.66 & \orangec .998 & \redc .0016         
               & \orangec 42.27 & \orangec .993 & \orangec .0065\\
        \hline       
    \end{tabular}
\end{center}
}
\caption{Quantitative comparison on a D-NeRF dataset}
\label{tab:apend1}
\end{table}

\begin{table}[]
{\small
\begin{center}
    \begin{tabular}{@{}l@{\;}|cc@{}c|c@{}c@{}c|c@{}c@{}c|c@{}c@{}c@{}}
        
        \hline
        Methods & \multicolumn{3}{c}{Bell} & \multicolumn{3}{c}{Sheet} & \multicolumn{3}{c}{Press} & \multicolumn{3}{c}{Basin}\\
        & PSNR $\uparrow$ & SSIM $\uparrow$ & LPIPS $\downarrow$ & PSNR $\uparrow$ & SSIM $\uparrow$ & LPIPS $\downarrow$ & PSNR $\uparrow$ & SSIM $\uparrow$ & LPIPS $\downarrow$ & PSNR $\uparrow$ & SSIM $\uparrow$ & LPIPS $\downarrow$
        \\\hline
        HyperNeRF & {\yellowc 24.0} & \orangec .884 & .159 
                  & 24.3 & \yellowc .874 & \yellowc .148 
                  & 25.4 & .873 & .164 
                  & 20.2 & .829 & .168
        \\
        NeRF-DS & 23.3 & .872 & \yellowc .134
                & \yellowc 25.7 & \redc .918 & \redc .115
                & \redc 26.4 & \redc .911 & \redc .123
                & 20.3 & \orangec .868 & \orangec .127
        \\
        TiNeuVox-B & 23.1 & \yellowc .876 & \redc .113
                   & 21.1 & .745 & .234
                   & 24.1 & .892 & \orangec .133
                   & \redc 20.7 & \redc .896 & \redc .105
        \\
        SC-GS & \orangec 25.1 & \redc .918 & \orangec .117
              & \redc 26.2 & \orangec .898 & \orangec .142
              & \redc 26.6 & \orangec.901 & \yellowc .135
              & 19.6 & \yellowc .846 & \yellowc .154 
        \\ 
        \our{} (our) & \redc 25.3 & .846 & .174 
               & \orangec 25.8 & .875 & .211
               & \yellowc 25.6 & .867 & .206
               & 19.8 & .788 & .217
        \\

        \hline
        Methods & \multicolumn{3}{c}{Cup} & \multicolumn{3}{c}{Sieve} & \multicolumn{3}{c}{Plant} & \multicolumn{3}{c}{Average} \\
        & PSNR $\uparrow$ & SSIM $\uparrow$ & LPIPS $\downarrow$ & PSNR $\uparrow$ & SSIM $\uparrow$ & LPIPS $\downarrow$ & PSNR $\uparrow$ & SSIM $\uparrow$ & LPIPS $\downarrow$ & PSNR $\uparrow$ & SSIM $\uparrow$ & LPIPS $\downarrow$
        \\\hline
        HyperNeRF & \orangec 20.5 & .705 & .318
                  & 25.0 & \yellowc .909 & \yellowc .129
                  & 18.1 & .714 & .359
                  & 22.5 & .827 & .206
        \\
        NeRF-DS & \redc 24.5 & \redc .916 & \orangec .118
                & \orangec 26.1 & \redc .935 & \redc .108
                & \redc 20.8 & \redc .867 & \orangec .164
                & \yellowc 23.9 & \redc .898 & \redc .127
        \\
        TiNeuVox-B & \orangec 20.5 & \yellowc .806 & \yellowc .182
                   & 20.1 & .822 & .205
                   & \orangec 20.6 & \orangec .863 & \redc .161
                   & 21.5 & .843 & \yellowc .162
        \\
        SC-GS & \redc 24.5 & \redc .916 & \redc .115
              & \yellowc 26.0 & \orangec .919 & \orangec .114
              & \yellowc 20.2 & \yellowc .837 & \yellowc .202
              & \redc 24.1 & \orangec .891 & \orangec .140
        \\
        \our{} (our) & \redc 24.5 & \orangec .874 & .185
               & \redc 26.5 & .881 & .159
               & \redc 20.8 & .815 & .234
               & \orangec 24.0 & \yellowc .849 & .198\\
        \hline       
    \end{tabular}
\end{center}}
\caption{Quantitative comparison on a NeRF-DS dataset}
\label{tab:apend2}
\end{table}

\begin{table}[]
{\small
\begin{center}
    \begin{tabular}{@{}l@{\;}|cc@{}c|c@{}c@{}c|c@{}c@{}c|c@{}c@{}c@{}}

        \hline
        Batch size & \multicolumn{3}{c}{Hook} & \multicolumn{3}{c}{Jumpingjacks} & \multicolumn{3}{c}{Trex} & \multicolumn{3}{c}{BouncingBalls}\\
        & PSNR $\uparrow$ & SSIM $\uparrow$ & LPIPS $\downarrow$ & PSNR $\uparrow$ & SSIM $\uparrow$ & LPIPS $\downarrow$ & PSNR $\uparrow$ & SSIM $\uparrow$ & LPIPS $\downarrow$ & PSNR $\uparrow$ & SSIM $\uparrow$ & LPIPS $\downarrow$
        \\\hline
         10 &\yellowc 37.02 &\yellowc.986 & \yellowc.0108 &  \redc 42.05 &  \redc .995 &  \redc .0049 &  \redc 40.88 &  \redc .996 &  \redc .0029 & \yellowc 39.19 & \orangec .988 & \yellowc .0132\\
        8 & \redc 38.13 & \redc .990 & \redc .0086
               &  \orangec 41.52 &  \redc .995 & \orangec .0052
               & \orangec 40.60 & \yellowc .994 &  \yellowc.0043
               &  \redc 41.49 &  \redc .993 & \orangec .0079\\
        4 & \orangec 37.68 & \orangec .989 &  \orangec .0105 & \yellowc40.44 & \yellowc.994 & \yellowc.0063 & \yellowc39.58 & \orangec .995 & \orangec .0047 & \orangec 39.76 & \redc .993 & \redc .0073\\

        \hline
        Batch size & \multicolumn{3}{c}{Hellwarrior} & \multicolumn{3}{c}{Mutant} & \multicolumn{3}{c}{Standup} & \multicolumn{3}{c}{Average} \\
        & PSNR $\uparrow$ & SSIM $\uparrow$ & LPIPS $\downarrow$ & PSNR $\uparrow$ & SSIM $\uparrow$ & LPIPS $\downarrow$ & PSNR $\uparrow$ & SSIM $\uparrow$ & LPIPS $\downarrow$ & PSNR $\uparrow$ & SSIM $\uparrow$ & LPIPS $\downarrow$
        \\\hline
            10 & \yellowc 40.02 & \yellowc .980 & \yellowc .0276 & \orangec 44.29 & \redc .997 & \orangec  .0027&  \redc 47.66 & \redc  .998 & \redc.0016  & \orangec 41.58 & \orangec  .990 & \yellowc .0091 \\
        8 &  \orangec 41.32 &  \redc .986 &  \orangec .0174
               &  \redc 44.38 &  \redc.997 & \redc .0026
               &  \orangec 47.17 &  \redc .998 & \orangec  .0018         
               &  \redc 42.08 & \redc .993 & \redc .0068\\
               4 & \redc 41.49 & \redc .986 & \redc .0173 & \yellowc43.32 & \yellowc.996 & \yellowc.0034 & \yellowc45.91 & \orangec .997 & \yellowc .0025  &\yellowc  41.16 & \redc .993 & \orangec .0074\\
        \hline       
    \end{tabular}
\end{center}
}
\caption{Batch study on a D-NeRF dataset}
\label{fig:apend3}
\end{table}

\begin{table}[]
{\small
\begin{center}
    \begin{tabular}{@{}l@{\;}|cc@{}c|c@{}c@{}c|c@{}c@{}c|c@{}c@{}c@{}}

        \hline
        Batch size & \multicolumn{3}{c}{Bell} & \multicolumn{3}{c}{Sheet} & \multicolumn{3}{c}{Press} & \multicolumn{3}{c}{Basin}\\
        & PSNR $\uparrow$ & SSIM $\uparrow$ & LPIPS $\downarrow$ & PSNR $\uparrow$ & SSIM $\uparrow$ & LPIPS $\downarrow$ & PSNR $\uparrow$ & SSIM $\uparrow$ & LPIPS $\downarrow$ & PSNR $\uparrow$ & SSIM $\uparrow$ & LPIPS $\downarrow$
        \\\hline
        8 & \orangec 25.1 & \redc .851 & \redc .165 & 24.8 & \yellowc .863 & .219 & \yellowc 25.2 & \yellowc .858 & \redc .205 & \orangec 19.7 & \redc.788 & \orangec .217
        \\
        4 & \yellowc 24.5 & .832 & \yellowc .190 & \yellowc 25.1 & \orangec .873 & \redc .202 & \orangec 25.3 & \orangec .860 & \yellowc .215 & \redc 19.8 & \yellowc .781 & .245 
        \\
        2 & \redc 25.3 & \orangec .846 & \orangec .174 & \redc 25.8 & \redc .875 & \orangec .211 & \redc 25.6 & \redc .867 & \orangec .206 & \yellowc 19.6 & \orangec .785 & \redc .207
        \\
        1 & \orangec 25.1 & \yellowc .844 & \yellowc .190 & \orangec 25.6 & \orangec .873 & \yellowc .216 & 24.3 & .842 & .273 & \orangec 19.7 & \orangec .785 & \yellowc .229
        \\

        \hline
        Batch size & \multicolumn{3}{c}{Cup} & \multicolumn{3}{c}{Sieve} & \multicolumn{3}{c}{Plant} & \multicolumn{3}{c}{Average} \\
        & PSNR $\uparrow$ & SSIM $\uparrow$ & LPIPS $\downarrow$ & PSNR $\uparrow$ & SSIM $\uparrow$ & LPIPS $\downarrow$ & PSNR $\uparrow$ & SSIM $\uparrow$ & LPIPS $\downarrow$ & PSNR $\uparrow$ & SSIM $\uparrow$ & LPIPS $\downarrow$
        \\\hline
        8 & 24.0 & \orangec .884 & \redc .163 & \redc 26.5 & \redc .881 & \redc .158 & 20.2 & \yellowc .812 & \redc .226 & \orangec 23.6 & \orangec .848 & \redc .193
        \\
        4 & \yellowc 24.2 & \redc .886 & \redc .163 & \yellowc 25.8 & \yellowc .871 & \orangec .167 & \redc 20.8 & \orangec .815 & \yellowc .234 & \orangec 23.6 & \yellowc .845 & \yellowc .202
        \\
        2 & \orangec 24.4 & \yellowc .883 & \orangec .169 & \orangec 26.0 & \orangec .875 & \yellowc .169 & \orangec 20.6 & \redc .817 & \orangec .232 & \redc 23.9 & \redc .849 & \orangec .195
        \\
        1 & \redc 24.5 & .874 & \yellowc .185 & \yellowc 25.8 & .864 & .195 & \yellowc 20.5 & .808 & .245 & \orangec 23.6 & .841 & .219\\
        \hline
    \end{tabular}
\end{center}
}
\caption{Batch study on a NeRF-DS dataset}
\label{fig:apend4}
\end{table}

\section{Extension of examples modification}

\begin{figure}[H]
    \centering
    \includegraphics[width=\textwidth, trim=0 100 0 0, clip]{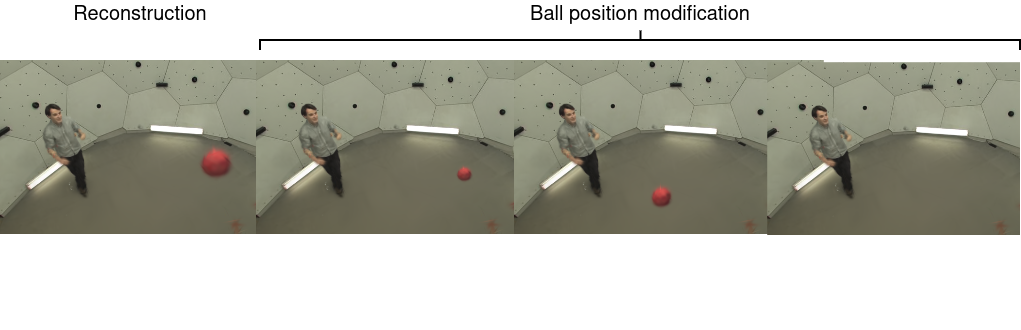}
    \caption{Example of a ball position modification on a $360^\circ$ scene from PanopticSports Datasets}
\label{fig:pilkirzucanie} 
\end{figure}

Below is an example of scene modification from the PanopticSports dataset (Fig. \ref{fig:pilkirzucanie}). The dynamics of the object has been stopped, so as to show the possibility of changing the position of the ball at a given time instant $t$. 

\begin{figure}[H]
    \centering
    \includegraphics[width=0.9\textwidth, trim=50 100 0 0, clip]{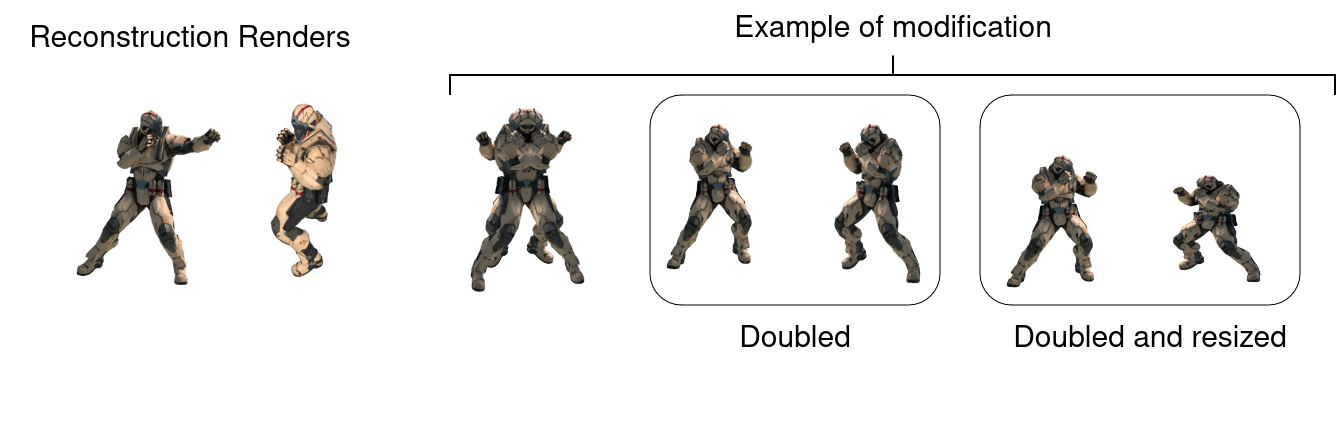}
    \caption{Example of animation by duplicating elements or changing their scale}
\label{fig:hool} 
\end{figure}

Due to the characteristics of Multi-Gaussians and the ease of their control -- we show that we can easily duplicate and/or scale selected objects (Fig. \ref{fig:hool}).  

\begin{figure}[H]
    \centering
    \includegraphics[width=0.9\textwidth, trim=0 0 0 0, clip]{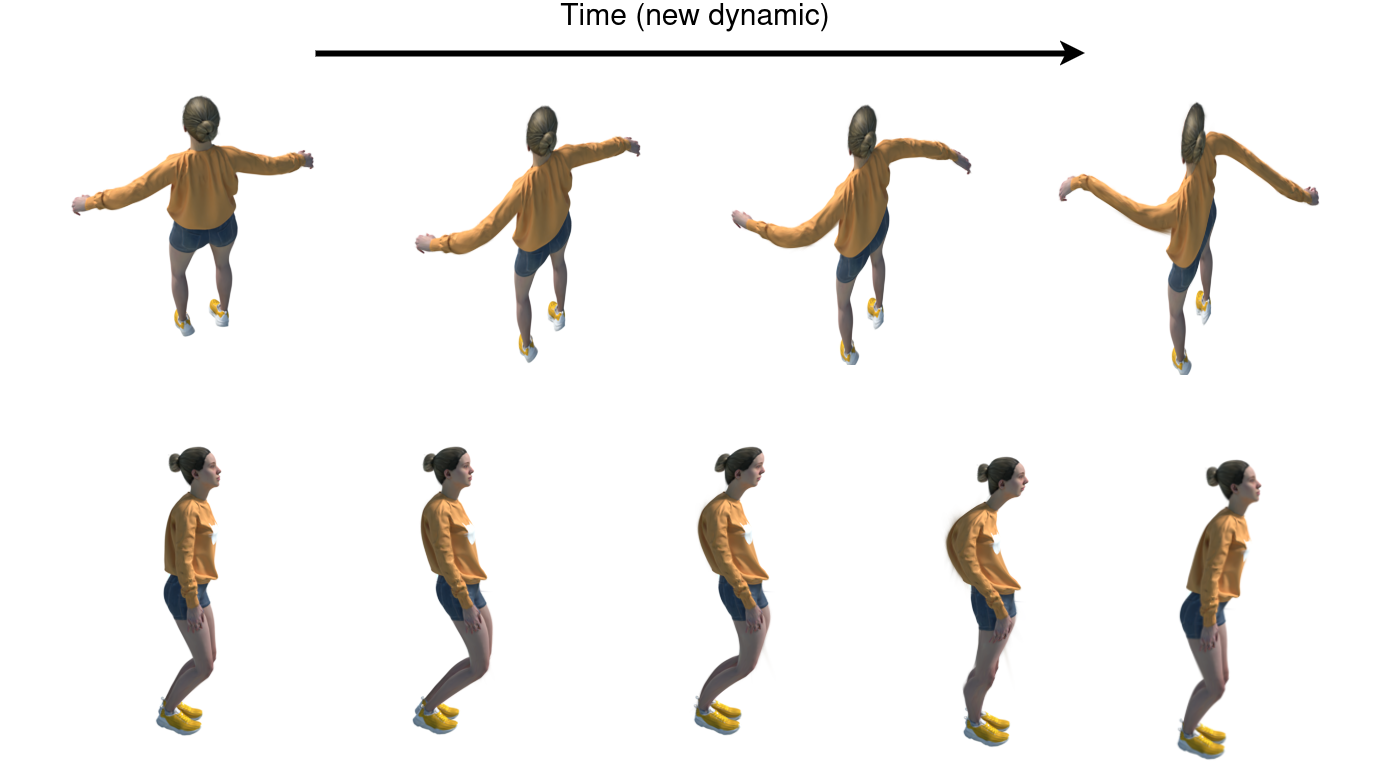}
    \caption{Example of third way of modification -- Transformation of space, allowing to give new dynamics.}
\label{fig:newdynamic} 
\end{figure}

Third method of modification allows to give new dynamics and fluidity to the object (Fig. \ref{fig:newdynamic}). We can see that even difficult edits like bending the back, changing the shape of the face, bending the hand are possible and the result is natural from the point of view of graphics, this is made possible by editing the scale and rotation of Gaussians in an explicitly defined way. 

\end{document}